\documentclass[a4paper,twoside]{article}

\usepackage{epsfig}
\usepackage{subcaption}
\usepackage{calc}
\usepackage{amssymb}
\usepackage{amstext}
\usepackage{amsmath}
\usepackage{amsthm}
\usepackage{multicol}
\usepackage{pslatex}
\usepackage{apalike}
\usepackage{algorithm2e}
\usepackage[bottom]{footmisc}
\usepackage{SCITEPRESS}     
\usepackage{booktabs}
\usepackage{multirow}
\usepackage{url}
\usepackage{tikz}

\usepackage{xcolor}

\definecolor{mhragdive}{HTML}{4C72B0}
\definecolor{mhragdiveroot}{HTML}{8172B2}
\definecolor{ragdive}{HTML}{DD8452}
\definecolor{clapnq}{HTML}{55A868}

\definecolor{mygold}{HTML}{E3C800}
\definecolor{mygoldborder}{HTML}{B09500}

\begin{document}

\title{MH-RAG-DIVE: Dynamic and Cost-Efficient Multi-Hop RAG Evaluation via Tree-Cached Dialogue Simulation}

\author{\authorname{Lorenz Brehme\sup{1,*}\orcidAuthor{0009-0009-4711-2564}, Daniel Hillebrand\sup{1,*}\orcidAuthor{0009-0000-8368-9138}, Thomas Ströhle\sup{1}\orcidAuthor{0000-0002-1954-6412} and Ruth Breu\sup{1}\orcidAuthor{0000-0001-7093-4341}}
\affiliation{\sup{1}University of Innsbruck, Innsbruck, Austria}
\affiliation{\sup{*}These authors contributed equally to this work.}
 \email{\{lorenz.brehme, daniel.hillebrand, thomas.stroehle, ruth.breu\}@uibk.ac.at}
}

\definecolor{mygold}{HTML}{E3C800}
\definecolor{mygoldborder}{HTML}{B09500}
\DeclareRobustCommand*\goldcircled[1]{%
  \tikz[baseline=(char.base)]{
    \node[
      shape=circle,
      draw=mygoldborder ,   
      fill=mygold,   
      text=black,   
      thick,
      inner sep=0.0pt,
      minimum size=10pt, 
    ] (char) {#1};}} 

\abstract{
Quality assessment is essential for determining whether changes to a Retrieval-Augmented Generation system improve or maintain its performance. However, existing evaluation approaches often rely on single-turn or single-hop interactions and therefore do not capture the complexity of multi-hop, multi-turn conversations. We introduce MH-RAG-DIVE, a dynamic interactive user simulation framework designed to evaluate RAG systems in more complex multi-hop and multi-turn settings. The framework extends RAG-DIVE with a Context Discoverer, which identifies documents that can be combined to generate multi-hop questions within an ongoing conversation. To address the high computational cost of dynamic user simulation, we further introduce a Conversation Cacher, which reuses turns from previous evaluation runs while preserving the dynamic nature of the evaluation.
We evaluate the framework across six distinct RAG configurations and compare its results with the original RAG-DIVE evaluation and the static, human-created CLAPNQ dataset. Our experiments show consistent performance trends across the evaluation approaches, while the more complex multi-hop and multi-turn setting exposes additional challenges for RAG systems. Furthermore, the Conversation Cacher substantially reduces token consumption, demonstrating its potential to make dynamic evaluation more cost-efficient. Finally, we apply MH-RAG-DIVE to an industrial RAG system with domain-specific data, demonstrating its applicability to real-world evaluation scenarios. Overall, our results show that the proposed approach provides a more complex and cost-efficient framework for evaluating RAG systems. All code and experimental data are publicly available in our GitHub repository (https://github.com/Daniel-Innsbruck/MH-RAG-DIVE).\\
\\
\textit{This preprint was accepted for presentation at the 19th International Conference on Agents and Artificial Intelligence (ICAART27).}
}

\onecolumn \maketitle \normalsize \setcounter{footnote}{0} \vfill

\section{Introduction}
Retrieval-Augmented Generation (RAG) was developed to address the limitations of Large Language Models (LLMs) in answering questions that require knowledge beyond their training data \cite{lewis2021retrievalaugmentedgenerationknowledgeintensivenlp}. The RAG pipeline, including document indexing, retrieval, and the integration of external knowledge into the generation process, offers numerous opportunities for customization \cite{kukreja_performance_2024,lyu_crud-rag_2025,Brehme_2025}.

Consequently, determining which parameter configurations yield the best performance for a specific domain becomes an important research challenge \cite{gao_retrieval-augmented_2024}.
Furthermore, these configurations require a standardized and reliable evaluation methodology to enable meaningful comparisons \cite{gao_retrieval-augmented_2024}. 

To obtain meaningful evaluations of RAG systems, the experimental setup should reproduce the characteristics of real-world information-seeking, both in the formulation of search queries and in human interactions with the system \cite{gao_retrieval-augmented_2024,brehme2025retrievalaugmentedgenerationindustryinterview}.
These interactions are typically simulated using a question–answer (QA) dataset, which can either be manually created by human experts or automatically generated by an LLM \cite{schimanski-etal-2024-climretrieve,cheng2024coralbenchmarkingmultiturnconversational}.
In practice, however, realistic information-seeking behaviour is difficult to capture because current evaluation methods often employ simplified settings, including single-turn interactions \cite{es2025ragasautomatedevaluationretrieval,lee2025gradegeneratingmultihopqa,tang2024multihopragbenchmarkingretrievalaugmentedgeneration} and questions that require information from only a single document \cite{chen2023benchmarkinglargelanguagemodels,liu2023recallbenchmarkllmsrobustness,brehme2026ragdivedynamicapproachmultiturn}. In contrast, real-world applications often require multi-hop reasoning, where answers must be synthesized from information distributed across multiple documents rather than retrieved from a single source \cite{lee2025gradegeneratingmultihopqa,tang2024multihopragbenchmarkingretrievalaugmentedgeneration}. Furthermore, users frequently engage in multi-turn conversations, in which the context established through previous interactions influences subsequent queries and responses \cite{brehme2026ragdivedynamicapproachmultiturn,zheng2023judgingllmasajudgemtbenchchatbot}.
Existing approaches, such as MultiHop-RAG \cite{tang2024multihopragbenchmarkingretrievalaugmentedgeneration}, which automatically generates multi-hop QA datasets, and RAG-DIVE \cite{brehme2026ragdivedynamicapproachmultiturn}, which dynamically evaluates RAG systems in a multi-turn setting, already address these requirements individually. However, they do not combine these capabilities. Specifically, existing approaches lack a framework for automatically evaluating RAG systems in a multi-hop, multi-turn setting.

To bridge this gap, we introduce MH-RAG-DIVE, which builds upon the RAG-DIVE framework and extends it by incorporating multi-hop queries. This is achieved through the introduction of the Context Discoverer, which identifies connected document chunks that are subsequently used to generate multi-hop questions. Additionally, we optimize the computational cost of dynamic interaction generation by introducing a caching algorithm that reduces unnecessary regeneration of questions and only creates new interactions when required.

\section{Related Work}

A comprehensive evaluation of a RAG system relies on an appropriate dataset that provides the information against which the system is evaluated \cite{brehme2026ragdivedynamicapproachmultiturn}. Early approaches to RAG evaluation predominantly used single-hop QA datasets, including SQuAD \cite{rajpurkar2016squad100000questionsmachine}. Single-hop queries can be answered using information retrieved from a single document or passage, requiring only one retrieval and reasoning step.

However, as RAG systems are increasingly deployed in knowledge-intensive domains, their evaluation must also account for multi-hop queries \cite{tang2024multihopragbenchmarkingretrievalaugmentedgeneration}. In contrast to single-hop queries, multi-hop queries require the model to retrieve and integrate information from multiple documents or passages before generating the correct answer. Individual retrieved contexts may appear irrelevant when considered in isolation, yet become essential when combined to support the required reasoning process \cite{park_dochop-qa_2025}. Consequently, evaluating RAG systems on multi-hop datasets provides a more realistic assessment of their retrieval capabilities and reasoning performance in complex, real-world applications \cite{tang2024multihopragbenchmarkingretrievalaugmentedgeneration,brehme2026evaluatingmultihopreasoningrag}.

Multi-hop RAG evaluation has traditionally been supported by specialised benchmarks, including HotpotQA and MuSiQue \cite{trivedi2022musiquemultihopquestionssinglehop,yang2018hotpotqadatasetdiverseexplainable}. HotpotQA established a benchmark setting in which answering a question requires the integration of evidence from multiple supporting documents, while distractor documents are included to assess the robustness of the retrieval process. MuSiQue builds upon this concept by composing multi-hop questions from simpler single-hop questions, thereby providing greater control over the structure of the underlying reasoning chains. Although such benchmarks have substantially contributed to the evaluation of multi-hop reasoning, their static nature introduces several challenges for contemporary RAG evaluation. In particular, publicly available datasets may be susceptible to data contamination resulting from their inclusion in pre-training corpora, potentially enabling LLMs to answer questions based on memorised knowledge rather than retrieved evidence \cite{singh2024evaluationdatacontaminationllms}. Moreover, fixed datasets provide limited opportunities to adapt evaluation scenarios to particular application domains or changing system requirements \cite{gan2025retrievalaugmentedgenerationevaluation}. To address these limitations, recent research has explored more flexible dataset construction methods, including domain-specific and synthetically generated evaluation data \cite{gan2025retrievalaugmentedgenerationevaluation,saadfalcon2024aresautomatedevaluationframework,tang2024multihopragbenchmarkingretrievalaugmentedgeneration}. Nevertheless, these methods typically generate evaluation instances independently of one another and therefore do not capture the evolving state of a conversational interaction, in which earlier retrieval outcomes and generated responses can affect subsequent dialogue states \cite{brehme2026ragdivedynamicapproachmultiturn}.

Despite the availability of numerous metrics for evaluating RAG systems, multi-hop capabilities are still frequently measured through system performance on multi-hop question answering benchmarks \cite{tang2024multihopragbenchmarkingretrievalaugmentedgeneration,brehme_automating_2026}. Such evaluations typically focus on the quality of the final answer using general answer-level metrics, leaving the intermediate reasoning and retrieval steps largely unassessed. Recent work has therefore begun to address the evaluation of these intermediate processes. For example, \cite{nguyen2024directevaluationchainofthoughtmultihop} introduce a framework that directly evaluates Chain-of-Thought reasoning in multi-hop question answering and show that a correct final answer does not necessarily imply a valid underlying reasoning process. 
From a complementary retrieval perspective, \cite{brehme2026evaluatingmultihopreasoningrag}  examine the assessment of retrieved information in multi-hop scenarios. They highlight that conventional retrieval metrics evaluate passages individually, even though the relevance of retrieved information may depend on the combination of multiple passages. To account for this, they propose a context-aware evaluation approach that assesses the retrieved context as a whole and reports higher agreement with human judgements than passage-level evaluation methods.

Whereas multi-hop evaluation has largely examined complex reasoning within individual queries, practical RAG applications commonly operate through ongoing conversational exchanges. This has motivated a shift towards multi-turn evaluation, in which the system is required not only to produce accurate responses but also to incorporate information from preceding interactions and maintain conversational context over time \cite{brehme2026ragdivedynamicapproachmultiturn}. Recent conversational benchmarks consequently include a range of interaction patterns, such as context-dependent follow-up questions, requests for clarification, and user corrections \cite{sirdeshmukh2025multichallengerealisticmultiturnconversation}. Nevertheless, the majority of existing multi-turn datasets remain static in their design \cite{cheng2024coralbenchmarkingmultiturnconversational,katsis2025mtragmultiturnconversationalbenchmark}. Their predefined conversational trajectories prevent subsequent queries from being influenced by the system's actual responses or preceding errors. As a result, such benchmarks provide only limited insight into the adaptive behaviour that RAG systems need to exhibit in realistic conversational settings \cite{brehme2026ragdivedynamicapproachmultiturn}.

RAG-DIVE addresses the shortcomings of static evaluation by replacing predefined test scenarios with dynamic LLM-to-LLM interactions \cite{brehme2026ragdivedynamicapproachmultiturn}. Its use of adaptive user personas enables a more realistic assessment of multi-turn RAG behaviour. However, the framework is designed around interactions with a single, isolated document \cite{brehme2026ragdivedynamicapproachmultiturn}. Consequently, it does not support the systematic evaluation of multi-hop reasoning across multiple documents within an evolving conversation. This limitation leaves room for evaluation frameworks that combine dynamic conversational interactions with the need to bridge information across multiple documents.

Although RAG-DIVE's LLM-driven dialogue generation allows conversational RAG systems to be evaluated under flexible and realistic interaction scenarios, generating these simulated dialogues introduces additional LLM inference overhead and consequently increases computational and financial costs \cite{brehme2026ragdivedynamicapproachmultiturn}. Cost efficiency is therefore an important factor in making LLM-based evaluation approaches scalable. One strategy for addressing the computational demands of repeated LLM inference is inference caching, which stores previously generated outputs for subsequent reuse \cite{liu2026semanticcachinglowcostllm}. Conventional caching mechanisms generally require an exact match between a new request and a previously cached request. Semantic caching relaxes this requirement by identifying requests with similar meanings and reusing the corresponding previously generated outputs \cite{liu2026semanticcachinglowcostllm}.

\section{MH-RAG-DIVE}
\label{Evalaution_strategy}
\begin{figure*}[ht]
    \centering
    \includegraphics[width=\linewidth]{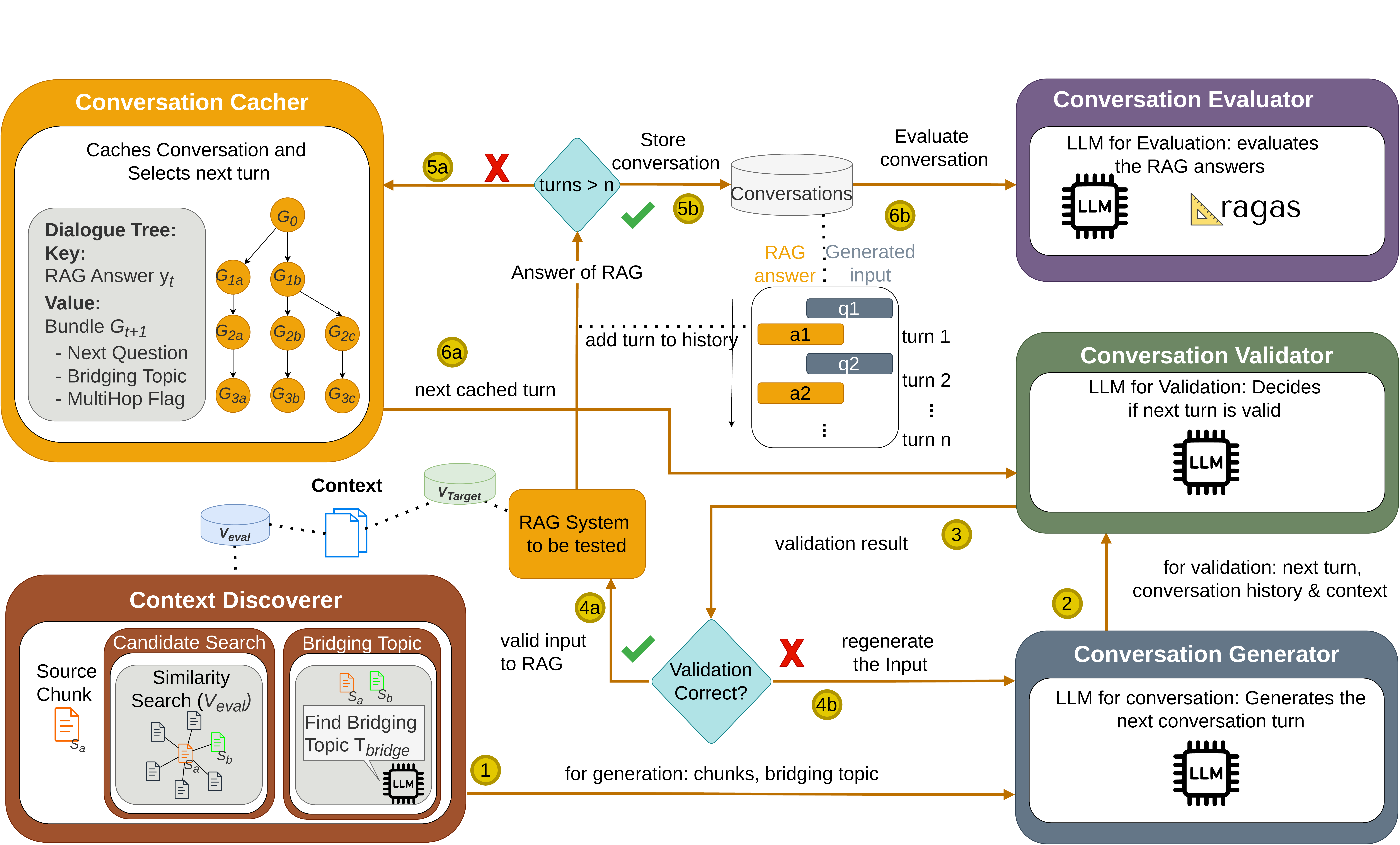}
    \caption{Illustration of MH-RAG-DIVE}
    \label{fig:RAG_DIVE}
    
\end{figure*}

To evaluate a RAG system in a multi-hop, multi-turn setting, we developed MH-RAG-DIVE. The framework extends the previously proposed RAG-DIVE approach \cite{brehme2026ragdivedynamicapproachmultiturn}.

MH-RAG-DIVE extends this architecture with two additional components and is composed of five core components (cf. Figure \ref{fig:RAG_DIVE}): (1) the Context Discoverer (CD), which identifies multiple relevant context chunks used to construct multi-hop questions requiring the integration of information from multiple sources; (2) the Conversation Generator (CG), which generates synthetic conversations; (3) the Conversation Validator (CV), which validates the generated conversations to ensure their quality and consistency; (4) the Conversation Cacher (CC), which reduces computational costs by caching previously generated conversations and regenerating only affected conversation turns instead of recreating entire conversations from scratch; and (5) the Conversation Evaluator (CE), which evaluates the responses generated by the RAG system.

\subsection{Context Discoverer}
The CD is the first component of the MH-RAG-DIVE framework. It initializes a dedicated vector database, referred to as the Evaluation Vector Database ($\mathcal{V}_{\text{eval}}$). This database is used to discover relevant context chunks for multi-hop question generation and to support context expansion throughout the course of a conversation.
\paragraph{Evaluation Vector Database}
By default, the evaluation set $\mathcal{V}_{\text{eval}}$ is derived directly from the raw document corpus. As part of the preprocessing pipeline, the documents are divided into overlapping chunks using recursive character-based splitting with fixed chunk lengths. The overlap between consecutive chunks helps maintain semantic continuity across chunk boundaries. Each resulting chunk $d_i$ is supplemented with descriptive metadata, such as document headers and creation dates, thereby preserving relevant contextual information from its source document. In addition, structural metadata is attached to the chunks to facilitate both multi-hop context selection and context management across multiple conversational turns.

Specifically, each chunk stores the identifier of its parent document, $P(d_i)$, which enables cross-document filtering for true multi-hop context sampling. Furthermore, each chunk is assigned its sequential position within the parent document, $Idx(d_i)$, together with the total number of chunks in that document, $N(d_i)$. These structural attributes enable deterministic context expansion throughout a conversation while preserving the original document order.

Finally, all chunks are embedded and stored in $\mathcal{V}_{\text{eval}}$, ensuring that the preprocessing pipeline remains independent of the downstream retrieval and generation architecture.
\paragraph{Two-Stage Multi-Hop Context Discovery}
To establish the multi-hop context for the initial query, a two-stage context discovery process is performed.
In the first stage, an arbitrary starting chunk, $d_A$, is sampled from $\mathcal{V}_{\text{eval}}$, and a candidate pool is constructed by retrieving its top-$k$ nearest neighbours. This retrieval is performed using a filtered semantic k-nearest neighbours search that enforces the constraint $P(d_j) \neq P(d_A)$, ensuring that all candidate chunks originate from different source documents and therefore enable cross-document reasoning.

The second stage selects a second chunk, $d_B$, with the objective of completing the initial multi-hop context. To identify a suitable candidate, chunks are sampled from the retrieved neighbourhood and assessed by an LLM for their thematic relationship to $d_A$. For each candidate, the LLM produces a potentially empty set of bridging topics, $T_{\text{bridge}}$, as well as a binary confidence score indicating whether the identified semantic relationship is adequate for generating a multi-hop question. Candidates receiving a confidence score of 0 are rejected, after which another candidate is sampled. This process continues until a suitable pair is found or all $k$ neighbours have been considered. If none of the candidates satisfies the required criteria, $d_A$ is discarded and the discovery process is restarted using a newly sampled initial chunk.

Once a valid pair has been identified, the resulting source set, $\mathcal{S}1 := \{d_A, d_B\}$, together with the corresponding bridging topics, $T_{\text{bridge}}$, is forwarded to the CG to generate the initial multi-hop query \goldcircled{1}.
\paragraph{Deterministic Context Expansion}
During the conversation, the CD incrementally expands the active context by adding adjacent chunks from the same source documents. This expansion is controlled by two optional trigger parameters, $\alpha$ and $\beta$ ($1 < \alpha, \beta \leq n$, where $n$ is the total number of dialogue turns), which specify the turns at which the contexts of $d_A$ and $d_B$ are expanded, respectively. Consequently, the context can be expanded either simultaneously ($\alpha = \beta$) or at different stages of the conversation. Before these trigger turns, the active context remains unchanged ($\mathcal{S}_t = \mathcal{S}_1$).

Context expansion follows a configurable traversal direction, $dir \in \{\text{above}, \text{below}\}$, which determines whether the preceding or succeeding chunk in the parent document is retrieved. If the selected direction reaches a document boundary, the traversal direction is automatically reversed. The retrieved adjacent chunk is then added to the active context, i.e., $\mathcal{S}_{t} = \mathcal{S}_{t-1} \cup \{d_{adj}\}$, allowing the CG to produce follow-up questions based on an expanded context while preserving the multi-hop nature of the dialogue.
\subsection{Conversation Generator}
The CG is responsible for generating the conversation. As in the original RAG-DIVE approach, it performs the same core tasks described by RAG-DIVE \cite{brehme2026ragdivedynamicapproachmultiturn}. To process the new multi-source input provided by the CD, the CG is prompted, following \cite{tang2024multihopragbenchmarkingretrievalaugmentedgeneration}, to formulate a question that requires integrating information from both context chunks based on the established semantic links. The CG's output schema is extended to include a final \textit{Bridging Topic}, which is synthesised from the candidate set $T_{\text{bridge}}$ and  the generated \textit{Query Type}. For this purpose, we used three types defined by \cite{tang2024multihopragbenchmarkingretrievalaugmentedgeneration}: inference, which refers to cases where information from one document must be retrieved and used as an intermediate step before the final search query can be formulated to identify the document containing the information required for the final answer; comparison, where two entities are compared; and temporal, where two items are related temporally. In the first turn, the question is required to involve multi-hop reasoning. In subsequent turns, multi-hop reasoning is optional. Therefore, only multi-hop questions in the first turn are classified as multi-hop questions, as subsequent turns do not introduce new documents but only extend the existing chunks. Consequently, we cannot reliably determine whether these follow-up questions genuinely require multi-hop reasoning. The CG produces a complete output bundle, denoted by $\mathcal{G}_1$, containing the \textit{RAG input}, \textit{standalone question}, \textit{query type}, \textit{ground-truth answer}, binary \textit{Multi-Hop Flag}, and synthesised \textit{bridging topic}. This bundle is then passed from the CG to the CV for subsequent processing \goldcircled{2}.

\subsection{Conversation Validator}
The CV validates the output of the CG. As in the original RAG-DIVE framework, it follows the validation procedure proposed by RAG-DIVE \goldcircled{3} \goldcircled{4} \cite{brehme2026ragdivedynamicapproachmultiturn}. In addition to the baseline pass/fail criteria, the CV verifies two further aspects for the initial turn: \textit{Query Type Alignment}, which determines whether the generated question accurately reflects the declared query type, and \textit{Multi-Hop Integrity}, which ensures that both context chunks are required to answer the question. Queries that can be answered using information from a single chunk are rejected.

\subsection{Conversation Cacher}
To avoid regenerating every conversation turn, we introduce the CC. This component caches each generated turn and determines whether regeneration is required.

The cache is organized as a tree structure. Initially, the source set $\mathcal{S}_1$ and the CG's initial output bundle $\mathcal{G}_1$ are stored as the root node. In subsequent runs, the framework retrieves these cached elements to instantly initialize the dialogue and establish a shared starting point.
Since RAG systems can produce variable responses, the cache tree grows dynamically by creating new branches for divergent RAG outputs. The CV and semantic similarity are used to navigate the cache and determine whether existing conversation paths can be reused or whether new branches must be generated.

At each turn $t$, a node corresponds to a specific dialogue state and maintains three primary components: the target system response $y_t$, the cached output bundle $\mathcal{G}_{t+1}$, and references to its child nodes. For terminal states, the latter is an empty set. The hyperparameters controlling the dialogue progression, $\alpha$, $\beta$, and $dir$, are stored together with the conversation nodes to maintain a consistent tree structure across different runs. Persisting these parameters ensures that the dynamic context expansion remains compatible with the cached dialogue trajectories.
In contrast, the active source set $\mathcal{S}_{t+1}$ is not stored as part of the cached state. Since context expansion is deterministically governed by the predefined triggers and remains synchronised with the turn index $t$.

Once the target RAG system delivers its current response $y_t$ \goldcircled{5a}, the CC computes the semantic similarity of $ y_t$ to all cached response keys $\tilde{Y}_t$ at the current node. 

The conversation continues from the cached response with the highest semantic similarity \goldcircled{6a}. The cached bundle $\tilde{\mathcal{G}}_{t+1}$ is then validated by the CV against the current conversation history $\mathcal{H}_t$ \goldcircled{3}. If the CV accepts the cached bundle, the corresponding turn is reused \goldcircled{4a}. If no suitable cached branch is available, the current turn is regenerated \goldcircled{4b}, and the resulting sequence is added as a new branch to the current node. The conversation is subsequently generated and cached through the final turn $n$, allowing the dialogue tree to expand incrementally as new system interactions are encountered.

\subsection{Conversation Evaluator}
The CE evaluates the RAG system's responses using the same evaluation criteria as the original RAG-DIVE framework \goldcircled{5b} \goldcircled{6b} \cite{brehme2026ragdivedynamicapproachmultiturn}. Specifically, it employs the four single-turn metrics—context recall, context precision, faithfulness, and correctness—as well as the two multi-turn metrics, forgetfulness and context retention. To provide a more fine-grained analysis, MH-RAG-DIVE extends this evaluation by filtering the single-turn metrics according to the \textit{Multi-Hop Flag}, thereby reporting performance separately for single-hop and multi-hop queries. This produces both an overall evaluation score and a detailed overview of the RAG system's performance across different retrieval and reasoning settings.
\section{Experimental Setup}
To validate MH-RAG-DIVE, we evaluate six different RAG setups on a common knowledge dataset, CLAPNQ \cite{rosenthal2024CLAPNQ}, and compare their performance against both the static CLAPNQ dataset and the original RAG-DIVE results. Additionally, we evaluate MH-RAG-DIVE in an industrial setting using a RAG system developed by Kufgem GmbH, a software company providing IT solutions for municipalities.

\paragraph{MH-RAG-DIVE}
For the knowledge base, we used the Wikipedia articles provided in the CLAPNQ dataset to construct $\mathcal{V}_{\text{eval}}$. We chunked the Wikipedia articles using a \textit{RecursiveCharacterTextSplitter} \cite{noauthor_langchain_nodate} with a chunk size of 3500 characters and an overlap of 525 characters, and used \textit{gemini-embedding-001} as the embedding model \cite{noauthor_embeddings_nodate}. This configuration served as the base $\mathcal{V}_{\text{eval}}$ for all experiments.
To run MH-RAG-DIVE, we used different LLMs for the individual components of the pipeline. For the CD, we used \textit{gemini-3.1-flash-lite} \cite{google2025gemini31flashlite}. For the CG, we used \textit{gemini-3.1-pro-preview} for the first turn \cite{google2025gemini31pro}, and \textit{gemini-3.1-flash-lite} in all subsequent turns \cite{google2025gemini31flashlite}. For the CV, we used \textit{gpt-5.6-luna} \cite{openai2025gpt5.6luna}, while \textit{gpt-5-nano} was used as for the CE \cite{openai2025gpt5nano}.

\paragraph{RAG Setups}
To run MH-RAG-DIVE, we configured six different RAG setups. We defined one base RAG system and varied a single parameter for each of the other configurations to investigate its effect on the evaluation results.

For the $\mathcal{V}_{\text{target}}$, we used the CLAPNQ dataset and chunked the documents with a chunk size of 500 characters and no overlap. This chunking configuration was kept fixed across all RAG setups.

For the embedding model, we used three different configurations. The base setup used \textit{gemini-embedding-001} \cite{noauthor_embeddings_nodate}. For comparison, we additionally used \textit{BAAI/bge-small-en-v1.5} \cite{noauthor_baaibge-small-en-v15_2026} and \textit{sentence-transformers/all-MiniLM-L6-v2} \cite{noauthor_sentence-transformersall-minilm-l6-v2_2024}.

We also varied the number of retrieved chunks. The base configuration retrieved five chunks per query, while additional configurations retrieved either one or eight chunks. Finally, we included an Agentic RAG setup that allows the system to perform multiple retrieval steps instead of relying on a single retrieval step, as in the base configuration.

Across all setups, we used \textit{gemma-4-31b-it} as the underlying LLM \cite{noauthor_modellkarte_nodate}. The RAG systems were implemented using LangChain \cite{noauthor_langchain_nodate}.

We conducted the experiments using 100 conversations with five turns each, resulting in a total of 500 QA pairs. Additionally, we performed a consistency check by running MH-RAG-DIVE ten times on the base RAG setup to assess the consistency and reproducibility of the results.

\paragraph{Human Validation}
To assess the quality of the generated questions, we sampled 100 turns and manually evaluated each turn and its corresponding conversation using the criteria proposed by \cite{lee2024multidocumentgroundedmultiturnsynthetic}. The evaluation was conducted independently by two individuals. At the turn level, we assessed correctness, answerability, and plausibility, while at the dialogue level, we evaluated diversity and coherence.
To additionally assess whether the multi-hop questions correctly capture the intended multi-hop structure, we introduced a new metric, multi-hop integrity. This metric measures whether all retrieved chunks are necessary to answer the question, thereby verifying that the question genuinely requires the intended multi-hop reasoning process.

\paragraph{Industrial Use Case}
To assess whether our approach generalizes to a domain-specific setting, we evaluated MH-RAG-DIVE on the RAG system of Kufgem GmbH, which is based on their k5\textbar Next platform \cite{k5nextK5NextClevere}. The k5\textbar Next platform provides digital administration solutions for municipalities. For this evaluation, we used the platform's documentation as the document base and constructed $\mathcal{V}_{\text{eval}}$ using the same parameters as in our other experiments.

We applied MH-RAG-DIVE to their RAG system and evaluated its performance using the synthetic dataset provided by them. We then compared the results obtained with MH-RAG-DIVE against the performance measured using the existing synthetic evaluation dataset and the RAG-DIVE baseline.

To facilitate reproducibility, all code and experimental data are publicly available in our GitHub repository \footnote{https://github.com/Daniel-Innsbruck/MH-RAG-DIVE}.

\section{Validation of MH-RAG-DIVE}
\label{valdiation_ragdive}
In this section, we evaluate the performance of MH-RAG-DIVE by comparing it with two baseline approaches, validating the results through human evaluation, and demonstrating its applicability in an industrial setting.

\subsection{Question Quality}
First, we validated the quality of the generated questions. The human evaluation showed that the question quality exceeded 90\% across all evaluated metrics, except for plausibility, where one reviewer rated only 89\% of the turns as plausible. 93\% of the questions were assessed as correct, meaning that both the ground-truth answer and the standalone question were generated correctly. Furthermore, 93\% of the questions were answerable using the provided chunks.

The multi-hop integrity metric showed that 95\% of the multi-hop questions in the first turn were genuine multi-hop questions that required combining information from both chunks to derive the answer, thereby providing the intended challenge for the RAG system. At the conversation level, 90\% of the conversations were assessed as coherent and diverse. Overall, these results demonstrate the high quality of both the generated questions and the resulting conversations.

\subsection{Conversation Cacher}
To validate the usefulness of the CC, we evaluated the acceptance rate of the CV for an existing turn. This acceptance rate is used as the cache hit rate. In the first turn, we observed a hit rate of 100\%, since the initial turn is reused for every new conversation, followed by 71.4\% in the second turn, 50\% in the third turn, 35\% in the fourth turn, and 23.7\% in the fifth turn. These results demonstrate how frequently a generated turn can be reused.

To evaluate the impact on token consumption, we compared the token usage of MH-RAG-DIVE with and without caching and also compared it with the original RAG-DIVE framework. Table \ref{tab:token_usage} shows that the CC reduces the number of CG tokens by 45\%. At the same time, it increases the number of CV tokens by approximately 11\%, resulting in an overall reduction of around 28\% in token consumption. Additionally, RAG-DIVE uses more than twice as many tokens, with a total of 233,330 tokens, indicating that MH-RAG-DIVE is less computationally expensive in terms of token usage.

Both approaches require the same number of tokens for the CE, which amounts to 480,000 tokens.
\begin{table}[ht]
    \centering
    \caption{Comparison of Average Token Usage Across MH-RAG-DIVE With and Without Caching and RAG-DIVE. MH-RD* denotes MH-RAG-DIVE without caching.}
    \begin{tabular}{lcccc}
    \toprule
        Approach & CG & CV & CD & Total  \\
         \midrule
        MH-RD & 30009 & 34618 & 0 & 64627 \\
        MH-RD* & 54801 & 31218 & 4671 & 90690 \\
        \midrule
        RAG-DIVE & 116731 & 116599 & 0 & 233330 \\
        \bottomrule
    \end{tabular} 
    \label{tab:token_usage}
\end{table}

\subsection{Consistency}

\begin{table}[ht]
\centering
\caption{Evaluation Consistency (n=10)}
\label{tab:consistency}
\begin{tabular}{lcccc}
\toprule
Metric & Mean & Std & Min & Max \\
\midrule
Context Retention  & 0.87    & 0.04    & 0.79    & 0.93    \\
Forgetfulness      & 0.30    & 0.05    & 0.21    & 0.41    \\
Correctness        & 0.63 & 0.02 & 0.60 & 0.66 \\
Faithfulness       & 0.71 & 0.01 & 0.7 & 0.73 \\
Context Precision  & 0.59 & 0.02 & 0.56 & 0.61 \\
Context Recall     & 0.67 & 0.01 & 0.64 & 0.69 \\
\bottomrule
\end{tabular}
\end{table}

To assess whether the results remain consistent when additional context is introduced, we ran the approach 10 times. We measured how often the CC reused a turn successfully and examined the variation in performance across runs for each setup.

Overall, we observed very low variance for the single-turn metrics, with a maximum standard deviation of 0.02 and a largest difference of 0.06 in total correctness. In contrast, the multi-turn metrics exhibited slightly higher variance, with values of 0.04 for context retention and 0.05 for forgetfulness. The sample size for these metrics is smaller, as they are based on 100 conversations rather than the 500 individual turns used for the single-turn evaluation. Looking at the multi-hop turns, which consist only of the initial turn, we observed similar performance, with a maximum variance of 0.02.

\subsection{Evaluation Sensitivity}

\begin{figure}[htbp]
    \centering

    \includegraphics[page=1,width=0.23\textwidth]{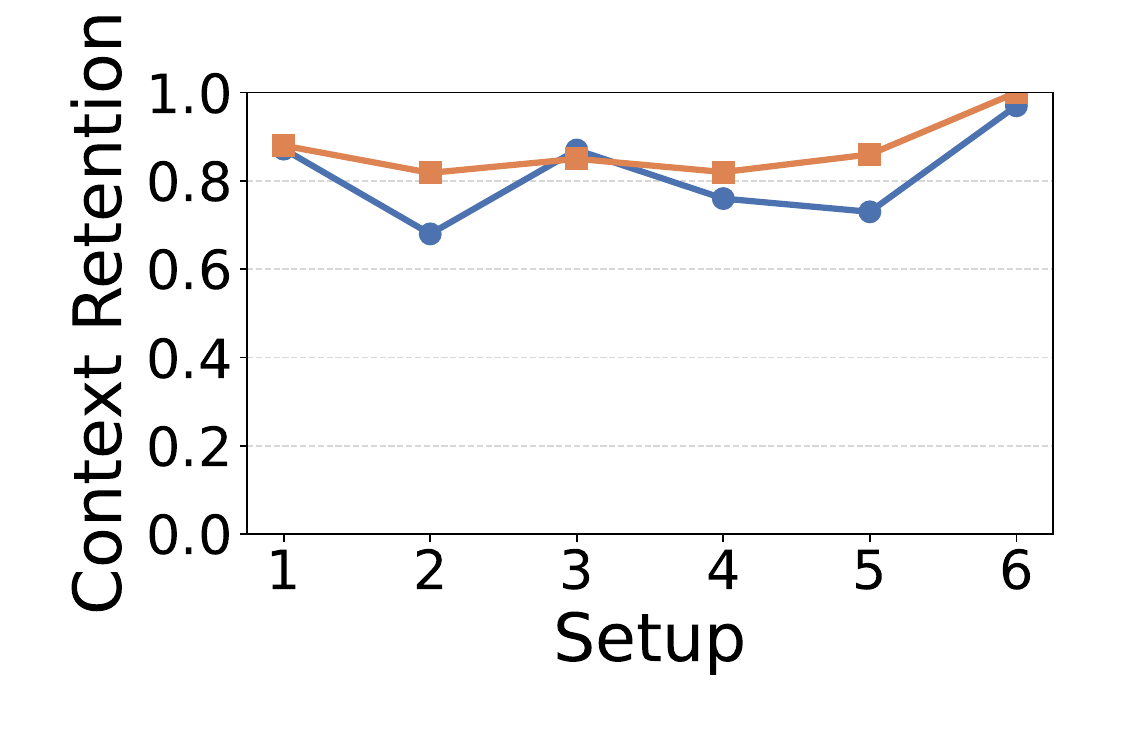}
    \includegraphics[page=1,width=0.23\textwidth]{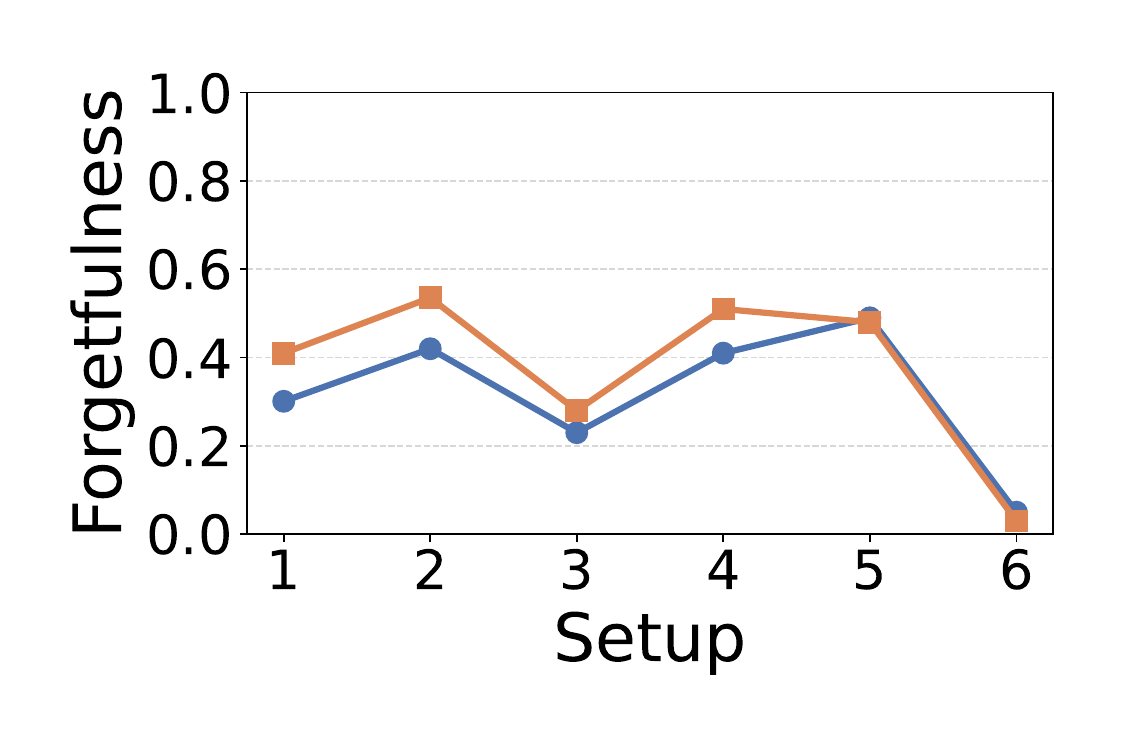}\\[0pt]

    \includegraphics[page=1,width=0.23\textwidth]{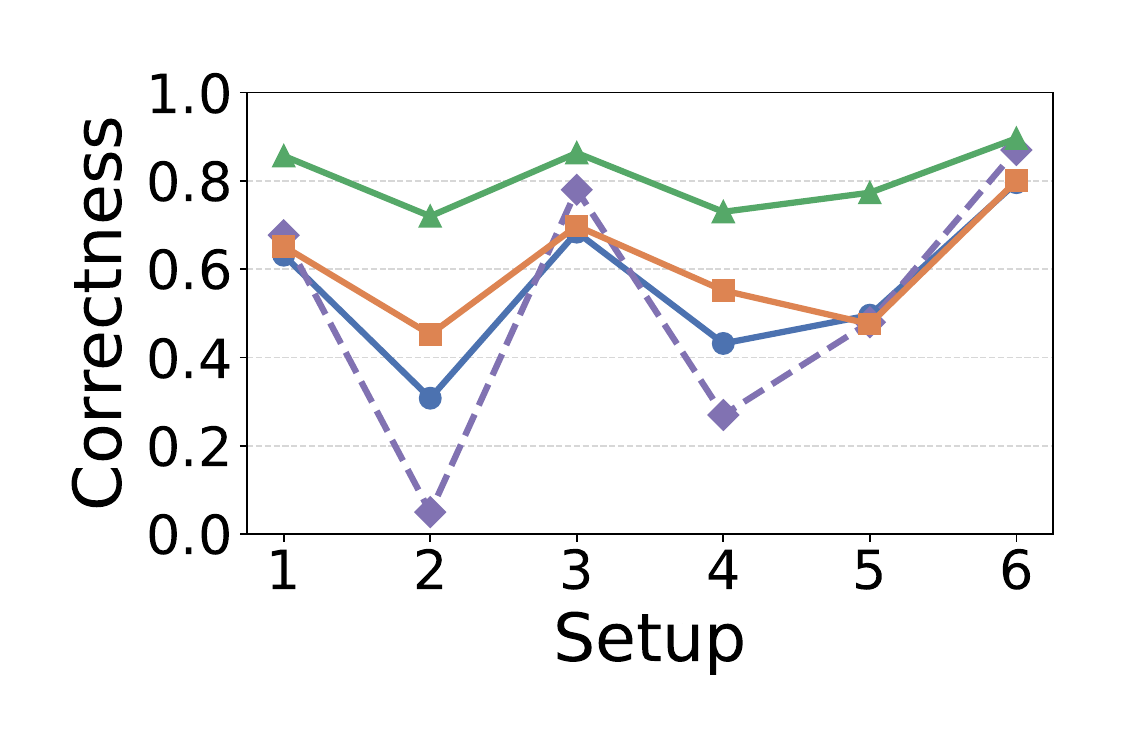}
    \includegraphics[page=1,width=0.23\textwidth]{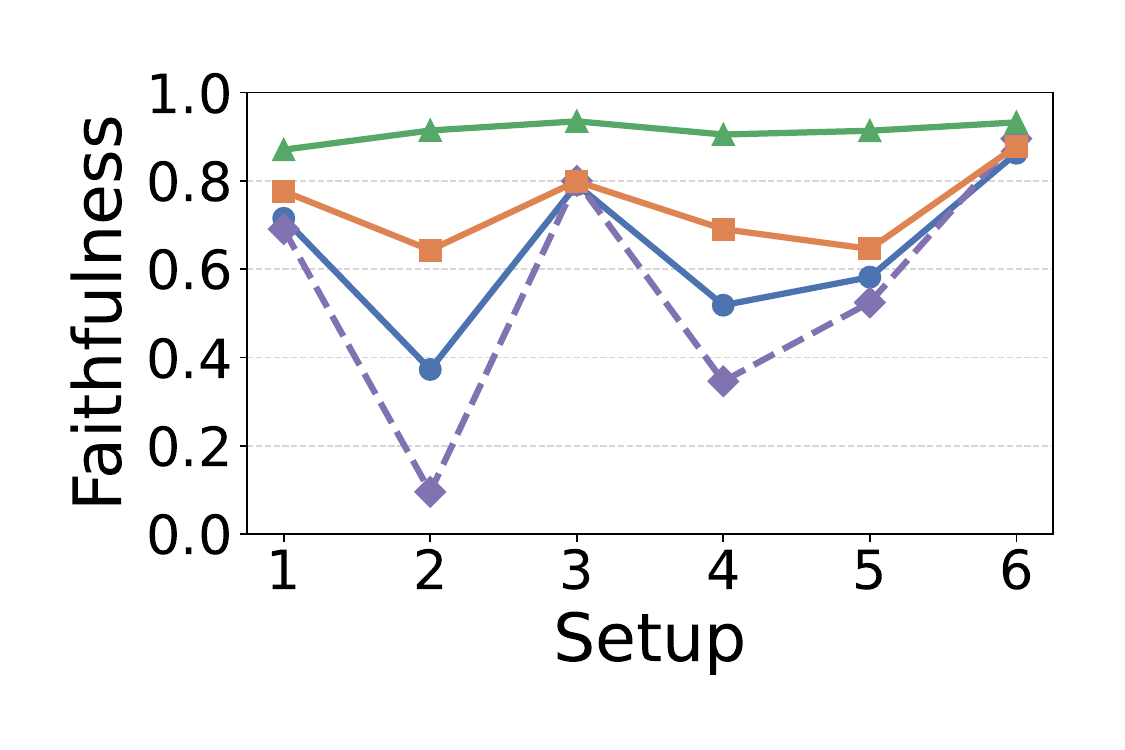}\\[0pt]

    \includegraphics[page=1,width=0.23\textwidth]{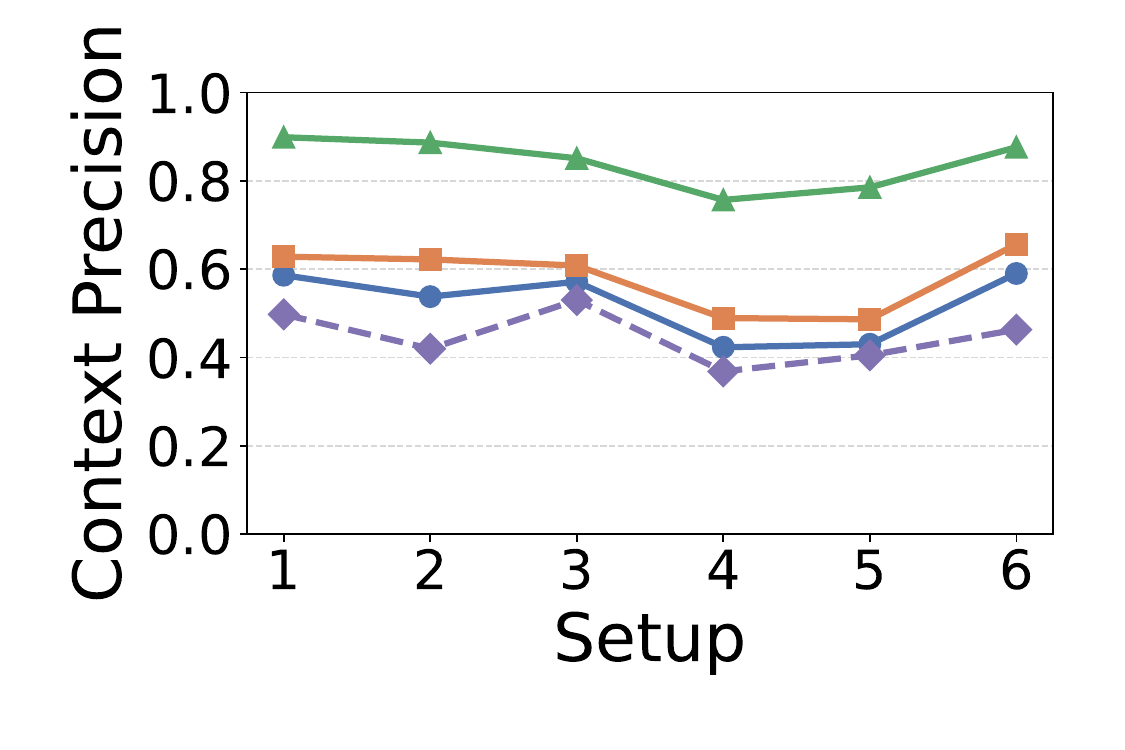}
    \includegraphics[page=1,width=0.23\textwidth]{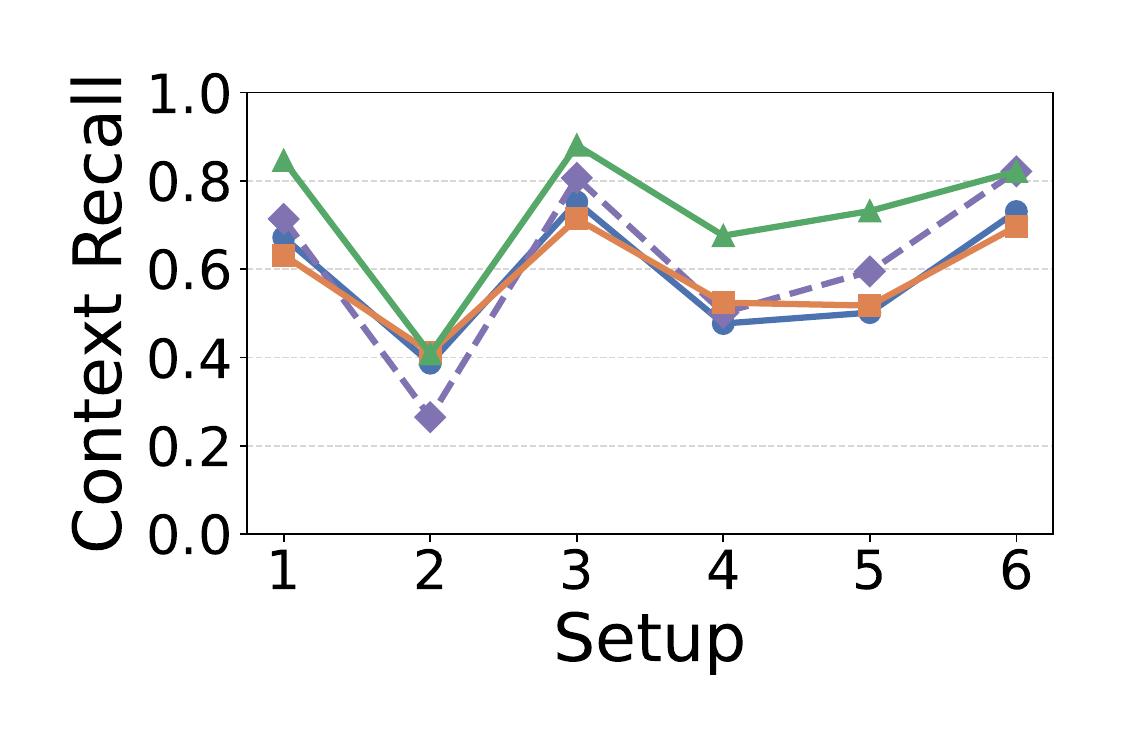}
    \caption{All metrics across six RAG configurations: (1) baseline, (2) one chunk, (3) eight chunks, (4) all-MiniLM embeddings, (5) BGE-small embeddings, and (6) agentic RAG. 
\textcolor{mhragdive}{\textbf{---$\bullet$--- MH-RAG-DIVE}} (all turns); 
\textcolor{mhragdiveroot}{\textbf{- - $\blacklozenge$ - - MH-RAG-DIVE}} (root turns only); 
\textcolor{ragdive}{\textbf{---$\blacksquare$--- RAG-DIVE}}; and 
\textcolor{clapnq}{\textbf{---$\blacktriangle$--- CLAPNQ}}.}
    \label{fig:six}
\end{figure}

To assess the sensitivity of our approach, we evaluated MH-RAG-DIVE across six different setups and compared its performance with the original RAG-DIVE and the human-generated questions from CLAPNQ (cf. Figure \ref{fig:six}). 

For the single-turn evaluation, we considered three question-generation settings.
Overall, the RAG system achieved the best performance on the CLAPNQ questions across all metrics, except for Context Recall in the one-chunk setup, where all approaches achieved comparable results. Comparing MH-RAG-DIVE with RAG-DIVE, the latter generally achieved higher Faithfulness and Context Precision. The two approaches performed comparably in Faithfulness for the Agentic and eight-chunk setups. For Correctness, both approaches achieved similar results in the Base, BGE-small, and Agentic setups, while RAG-DIVE performed better in the remaining setups. Context Recall remained comparable between MH-RAG-DIVE and RAG-DIVE across all evaluated setups.

Looking at the performance trends, all three approaches exhibit broadly similar patterns across the evaluated setups. Correctness is highest for the Agentic setup, followed by the eight-chunk configuration, while the one-chunk configuration consistently achieves the lowest correctness. For MH-RAG-DIVE and CLAPNQ, the one-chunk setup is followed by the All-MiniLM and then the BGE-small configurations. In contrast, for RAG-DIVE, BGE-small achieved better performance than All-MiniLM.

Faithfulness follows a similar trend. For Context Precision, however, the two embedding-based configurations generally perform worse than the one-chunk setup. For both CLAPNQ and RAG-DIVE, increasing the context to eight chunks does not improve Context Precision compared to the one-chunk configuration. In contrast, MH-RAG-DIVE benefits from the larger context, achieving higher Context Precision with eight chunks than with one chunk.

The trends for Context Recall largely resemble those observed for Correctness, with one notable difference: the eight-chunk configuration achieves higher Context Recall than the Agentic setup.

Overall, MH-RAG-DIVE exhibits larger performance differences across the evaluated setups than RAG-DIVE. For example, Correctness in the one-chunk setup is approximately 15 percentage points lower for MH-RAG-DIVE than for RAG-DIVE, whereas both approaches achieve comparable performance with eight chunks. In contrast, the CLAPNQ results remain relatively consistent across setups, with little variation in most metrics, except for Context Recall, which shows more pronounced differences.

The multi-turn metrics are available only for MH-RAG-DIVE and RAG-DIVE, as CLAPNQ consists exclusively of single-turn interactions. For Context Retention, RAG-DIVE achieves equal or better performance than MH-RAG-DIVE across all evaluated setups. In contrast, MH-RAG-DIVE performs equally well or better than RAG-DIVE on Forgetfulness. The overall performance trends are similar across the two approaches, with the main differences again observed between the two embedding-based configurations, consistent with the single-turn results.

\subsubsection{MultiHop Questions}

In addition to the overall scores, MH-RAG-DIVE provides a separate score for multi-hop questions by labeling the initial turn as a multi-hop question. This enables a more fine-grained analysis of RAG performance specifically on multi-hop questions (cf. Figure \ref{fig:six}). The results indicate that the RAG systems generally struggle more with multi-hop questions, as reflected by their lower scores. This effect is particularly pronounced for the one-chunk configuration and the setup using the BGE-small embedding model.
For the remaining setups, the differences are generally smaller, with multi-hop questions yielding mostly lower or comparable performance. The only notable exception is Context Precision, where the multi-hop configurations achieve higher scores in the Agentic and eight-chunk setups.
\subsection{Industrial Use Case}

\begin{table}[ht]
\centering
\caption{Comparison of the original setup (Ori) with RAG-DIVE (RD) and MH-RAG-DIVE (MH-RD) for the industrial use case.}
\label{tab:industrial_use_case}
\setlength{\tabcolsep}{6pt}
\renewcommand{\arraystretch}{1.0}
\small
\begin{tabular}{lccc}
\toprule
Metric
& MH-RD
& RD
& Ori \\
\midrule
Context Retention
& 1.0 & 0.975 & -- \\

Forgetfulness
& 0.00 & 0.10 & -- \\

Correctness
& 0.65 & 0.84 & 0.88 \\

Faithfulness
& 0.61 & 0.66 & 0.71 \\

Context Precision
& 0.63 & 0.87 & 0.95 \\

Context Recall
& 0.68 & 0.85 & 0.73 \\
\bottomrule
\end{tabular}
\end{table}

To evaluate MH-RAG-DIVE in an industrial setting and assess whether the approach generalizes beyond an artificial benchmark, we evaluated three approaches on an industrial RAG system provided by Kufgem GmbH. 
We compared RAG-DIVE and MH-RAG-DIVE against the company's existing QA set, which is currently used for evaluating and optimizing their RAG system.

Using RAG-DIVE and MH-RAG-DIVE enables evaluation in a multi-turn setting and additionally provides two multi-turn metrics: Context Retention and Forgetfulness. The remaining single-turn metrics are available for all three approaches. Overall, the RAG system showed the lowest performance on MH-RAG-DIVE, followed by RAG-DIVE, while achieving the highest performance on the existing QA set, for which the system has already been evaluated and optimized. The only exception was Context Recall, where the RAG system performed better with RAG-DIVE than with the existing QA set.

\section{Discussion}
Our experiments demonstrate the applicability of our proposed evaluation approach in both an artificial setup based on Wikipedia data and a domain-specific setting using technical documentation. In both settings, it provides a more challenging evaluation of RAG systems, reflected in lower scores compared to the RAG-DIVE baseline and simple single-turn QA pairs. This suggests that the multi-hop formulation can expose weaknesses that may remain hidden when evaluating RAG systems using simpler, single-turn questions.

Despite the increased difficulty, the consistency checks show that MH-RAG-DIVE produces stable results across repeated runs, indicating that it can provide a reliable evaluation. This consistency is supported by the caching and validation mechanisms of the CV. Furthermore, comparisons across the three evaluated approaches demonstrate consistent performance patterns across different experimental setups.

In general, RAG systems struggle more with MH-RAG-DIVE because the multi-hop questions require the retrieval of multiple relevant pieces of information. This effect is particularly visible in the one- and eight-chunk configurations. In the eight-chunk setup, however, the performance difference compared to the RAG-DIVE evaluation is less pronounced, as the multi-hop questions require several chunks to be retrieved to obtain the correct answer. In contrast, the other evaluation settings contain no multi-hop questions, meaning that a single correctly retrieved chunk can often be sufficient to answer the question. Consequently, the difference between the one- and eight-chunk configurations is more pronounced for MH-RAG-DIVE.
Overall, RAG-DIVE exhibits smaller performance differences between configurations than MH-RAG-DIVE. We attribute this to the lower difficulty of the questions used by RAG-DIVE: simpler questions can often be answered successfully even with less effective retrieval configurations. In contrast, the more challenging multi-hop questions in MH-RAG-DIVE amplify differences between retrieval setups, making the strengths and weaknesses of each configuration more apparent.

Additionally, the eight-chunk configuration achieves higher Context Recall than the Agentic setup. With eight chunks, a larger number of potentially relevant chunks is retrieved deterministically, whereas the Agentic setup relies on the agent to determine how many chunks to retrieve. If the agent retrieves fewer chunks, relevant information may be missed, resulting in lower recall.

An observation concerns the choice of embedding model. While all-MiniLM performed worst among the evaluated embedding configurations for MH-RAG-DIVE and CLAPNQ, the BGE-small model achieved better performance for RAG-DIVE. This difference in performance trends warrants further investigation to understand why the effectiveness of the embedding models varies across the evaluation approaches.

In addition to improving question quality and increasing the difficulty of the evaluation, we introduce the CC, which reduces token usage by reusing previously generated questions. This reduction in token consumption can lower computational and energy costs. Our experiments demonstrate that the CC successfully reuses previously generated questions and quantify the resulting savings compared to the original approach.
Despite the reduction in token usage achieved by the Conversation Cacher, the majority of the tokens are still consumed by the CE, which currently relies on the RAGAS framework. Future work could therefore investigate ways to reduce the computational and token costs of this component. Additionally, evaluating the CE using self-hosted models could further reduce reliance on external APIs and associated costs, while also providing insights into the performance and feasibility of locally deployed evaluation models.

One potential limitation of our approach is retriever bias arising from the architecture of the CD, which retrieves chunks based on semantic similarity. If the same retriever is used by the evaluated RAG system, this could favor systems that employ similar retrieval mechanisms and potentially lead to artificially similar performance. To investigate this possibility, we evaluated three different RAG configurations using different embedding models and examined whether they exhibited similar performance patterns. The observed trends were similar to those of the human-generated CLAPNQ questions, whereas RAG-DIVE showed different trends, with BGE-small performing worst rather than All-MiniLM. These differences suggest that further investigation is required to determine whether retriever bias affects the evaluation. Therefore, the potential influence of retriever bias should be considered when interpreting the results and determining which RAG configuration performs best. Future work will investigate this effect in greater detail.

A second potential limitation is LLM circularity, as an LLM is used to validate questions generated by an LLM. This setup could introduce biases if the evaluator favors outputs produced by models with similar characteristics. To mitigate this risk, we used different LLM architectures for question generation and validation and additionally conducted a human evaluation to verify the quality of the generated conversations. Nevertheless, users of the approach should remain aware of the potential limitations of LLM-as-a-judge evaluation and avoid overfitting RAG systems to the evaluation model. Additional evaluation is therefore required before drawing definitive conclusions about the best-performing RAG configuration. In practice, this could include human-in-the-loop evaluation or feedback from real users.

\section{Conclusion}

This paper introduces an extension of the RAG-DIVE evaluation approach by incorporating the CD to generate multi-hop questions, enabling a more complex and comprehensive evaluation. Additionally, it introduces the CC to reduce token usage and, consequently, the cost of the evaluation process.

Our experiments demonstrate the effectiveness of the proposed approach compared with the RAG-DIVE baseline and show that the CC can substantially reduce token consumption. Overall, this extension provides a more complex and cost-efficient evaluation and demonstrates its applicability in an industrial use case involving domain-specific data.

\section*{\uppercase{Acknowledgements}}
We would like to thank Kufgem GmbH for providing access to their RAG system and the associated data, which enabled us to validate our approach using a real-world, domain-specific application.

During the preparation of this paper, we used ChatGPT and Grammarly for language editing, grammar, and spelling checks. GitHub Copilot and Claude Code were used to assist with software development and coding-related tasks. All generated content and code were reviewed and verified by the authors.

\bibliographystyle{apalike}
{\small 
\bibliography{example}}

@misc{lewis2021retrievalaugmentedgenerationknowledgeintensivenlp,
      title={Retrieval-Augmented Generation for Knowledge-Intensive NLP Tasks}, 
      author={Patrick Lewis and Ethan Perez and Aleksandra Piktus and Fabio Petroni and Vladimir Karpukhin and Naman Goyal and Heinrich Küttler and Mike Lewis and Wen-tau Yih and Tim Rocktäschel and Sebastian Riedel and Douwe Kiela},
      year={2021},
      eprint={2005.11401},
      archivePrefix={arXiv},
      primaryClass={cs.CL},
      url={https://arxiv.org/abs/2005.11401}, 
}

@inproceedings{Brehme_2025,
   title={Can LLMs be Trusted for Evaluating RAG Systems? A Survey of Methods and Datasets},
   url={http://dx.doi.org/10.1109/SDS66131.2025.00010},
   DOI={10.1109/sds66131.2025.00010},
   booktitle={2025 IEEE Swiss Conference on Data Science (SDS)},
   publisher={IEEE},
   author={Brehme, Lorenz and Ströhle, Thomas and Breu, Ruth},
   year={2025},
   month=jun, pages={16–23} }

@misc{lee2025gradegeneratingmultihopqa,
	title={GRADE: Generating multi-hop QA and fine-gRAined Difficulty matrix for RAG Evaluation}, 
	author={Jeongsoo Lee and Daeyong Kwon and Kyohoon Jin},
	year={2025},
	eprint={2508.16994},
	archivePrefix={arXiv},
	primaryClass={cs.CL},
	url={https://arxiv.org/abs/2508.16994}, 
}

@misc{es2025ragasautomatedevaluationretrieval,
	title={Ragas: Automated Evaluation of Retrieval Augmented Generation}, 
	author={Shahul Es and Jithin James and Luis Espinosa-Anke and Steven Schockaert},
	year={2025},
	eprint={2309.15217},
	archivePrefix={arXiv},
	primaryClass={cs.CL},
	url={https://arxiv.org/abs/2309.15217}, 
}

@misc{tang2024multihopragbenchmarkingretrievalaugmentedgeneration,
	title={MultiHop-RAG: Benchmarking Retrieval-Augmented Generation for Multi-Hop Queries}, 
	author={Yixuan Tang and Yi Yang},
	year={2024},
	eprint={2401.15391},
	archivePrefix={arXiv},
	primaryClass={cs.CL},
	url={https://arxiv.org/abs/2401.15391}, 
}

@conference{brehme2025retrievalaugmentedgenerationindustryinterview,
author={Lorenz Brehme and Benedikt Dornauer and Thomas Ströhle and Maximilian Ehrhart and Ruth Breu},
title={Retrieval-Augmented Generation in Industry: An Interview Study on Use Cases, Requirements, Challenges, and Evaluation},
booktitle={Proceedings of the 17th International Joint Conference on Knowledge Discovery, Knowledge Engineering and Knowledge Management - KDIR},
year={2025},
pages={110-122},
publisher={SciTePress},
organization={INSTICC},
doi={10.5220/0013739500004000},
isbn={},
issn={2184-3228},
}

@misc{chen2023benchmarkinglargelanguagemodels,
	title={Benchmarking Large Language Models in Retrieval-Augmented Generation}, 
	author={Jiawei Chen and Hongyu Lin and Xianpei Han and Le Sun},
	year={2023},
	eprint={2309.01431},
	archivePrefix={arXiv},
	primaryClass={cs.CL},
	url={https://arxiv.org/abs/2309.01431}, 
}

@misc{liu2023recallbenchmarkllmsrobustness,
	title={RECALL: A Benchmark for LLMs Robustness against External Counterfactual Knowledge}, 
	author={Yi Liu and Lianzhe Huang and Shicheng Li and Sishuo Chen and Hao Zhou and Fandong Meng and Jie Zhou and Xu Sun},
	year={2023},
	eprint={2311.08147},
	archivePrefix={arXiv},
	primaryClass={cs.CL},
	url={https://arxiv.org/abs/2311.08147}, 
}

@misc{saadfalcon2024aresautomatedevaluationframework,
	title={ARES: An Automated Evaluation Framework for Retrieval-Augmented Generation Systems}, 
	author={Jon Saad-Falcon and Omar Khattab and Christopher Potts and Matei Zaharia},
	year={2024},
	eprint={2311.09476},
	archivePrefix={arXiv},
	primaryClass={cs.CL},
	url={https://arxiv.org/abs/2311.09476}, 
}

@misc{cheng2024coralbenchmarkingmultiturnconversational,
	title={CORAL: Benchmarking Multi-turn Conversational Retrieval-Augmentation Generation}, 
	author={Yiruo Cheng and Kelong Mao and Ziliang Zhao and Guanting Dong and Hongjin Qian and Yongkang Wu and Tetsuya Sakai and Ji-Rong Wen and Zhicheng Dou},
	year={2024},
	eprint={2410.23090},
	archivePrefix={arXiv},
	primaryClass={cs.IR},
	url={https://arxiv.org/abs/2410.23090}, 
}

@inproceedings{brehme2026ragdivedynamicapproachmultiturn,
author = {Brehme, Lorenz and Dornauer, Benedikt and B{\"o}ttcher, Jan-Henrik and Schmid, Klaus and Racasan, Mircea-Cristian and Breu, Ruth},
title = {RAG-DIVE: A Dynamic Approach for Multi-Turn Dialogue Evaluation in Retrieval-Augmented Generation},
year = {2026},
isbn = {9798400724756},
publisher = {Association for Computing Machinery},
address = {New York, NY, USA},
url = {https://doi.org/10.1145/3793653.3793766},
doi = {10.1145/3793653.3793766},
booktitle = {Proceedings of the IEEE/ACM 5th International Conference on AI Engineering - Software Engineering for AI},
pages = {74–84},
numpages = {11},
location = {
},
series = {CAIN '26}
}

@misc{sirdeshmukh2025multichallengerealisticmultiturnconversation,
	title={MultiChallenge: A Realistic Multi-Turn Conversation Evaluation Benchmark Challenging to Frontier LLMs}, 
	author={Ved Sirdeshmukh and Kaustubh Deshpande and Johannes Mols and Lifeng Jin and Ed-Yeremai Cardona and Dean Lee and Jeremy Kritz and Willow Primack and Summer Yue and Chen Xing},
	year={2025},
	eprint={2501.17399},
	archivePrefix={arXiv},
	primaryClass={cs.CL},
	url={https://arxiv.org/abs/2501.17399}, 
}

@misc{yang2018hotpotqadatasetdiverseexplainable,
      title={HotpotQA: A Dataset for Diverse, Explainable Multi-hop Question Answering}, 
      author={Zhilin Yang and Peng Qi and Saizheng Zhang and Yoshua Bengio and William W. Cohen and Ruslan Salakhutdinov and Christopher D. Manning},
      year={2018},
      eprint={1809.09600},
      archivePrefix={arXiv},
      primaryClass={cs.CL},
      url={https://arxiv.org/abs/1809.09600}, 
}

@misc{trivedi2022musiquemultihopquestionssinglehop,
	title={MuSiQue: Multihop Questions via Single-hop Question Composition}, 
	author={Harsh Trivedi and Niranjan Balasubramanian and Tushar Khot and Ashish Sabharwal},
	year={2022},
	eprint={2108.00573},
	archivePrefix={arXiv},
	primaryClass={cs.CL},
	url={https://arxiv.org/abs/2108.00573}, 
}

@misc{gan2025retrievalaugmentedgenerationevaluation,
	title={Retrieval Augmented Generation Evaluation in the Era of Large Language Models: A Comprehensive Survey}, 
	author={Aoran Gan and Hao Yu and Kai Zhang and Qi Liu and Wenyu Yan and Zhenya Huang and Shiwei Tong and Guoping Hu},
	year={2025},
	eprint={2504.14891},
	archivePrefix={arXiv},
	primaryClass={cs.CL},
	url={https://arxiv.org/abs/2504.14891}, 
}

@misc{zheng2023judgingllmasajudgemtbenchchatbot,
	title={Judging LLM-as-a-Judge with MT-Bench and Chatbot Arena}, 
	author={Lianmin Zheng and Wei-Lin Chiang and Ying Sheng and Siyuan Zhuang and Zhanghao Wu and Yonghao Zhuang and Zi Lin and Zhuohan Li and Dacheng Li and Eric P. Xing and Hao Zhang and Joseph E. Gonzalez and Ion Stoica},
	year={2023},
	eprint={2306.05685},
	archivePrefix={arXiv},
	primaryClass={cs.CL},
	url={https://arxiv.org/abs/2306.05685}, 
}

@manual{google2025gemini31flashlite,
	title        = {Gemini 3.1 Flash Lite Documentation and Model Specifications},
	author       = {{Google Cloud}},
	year         = {2026},
	url          = {https://docs.cloud.google.com/gemini-enterprise-agent-platform/models/gemini/3-1-flash-lite},
	note         = {Accessed: 2026-08-20},
	organization = {Google LLC}
}

@manual{google2025gemini31pro,
	title        = {Gemini 3.1 Pro Documentation and Model Specifications},
	author       = {{Google Cloud}},
	year         = {2026},
	url          = {https://docs.cloud.google.com/gemini-enterprise-agent-platform/models/gemini/3-1-pro?hl=de},
	note         = {Accessed: 2026-08-20},
	organization = {Google LLC}
}

@manual{openai2025gpt5nano,
	title        = {OpenAI Models Documentation: GPT-5-nano Specifications},
	author       = {{OpenAI}},
	year         = {2025},
	url          = {https://developers.openai.com/api/docs/models/gpt-5-nano},
	note         = {Accessed: 2026-09-21},
	organization = {OpenAI}
}

@manual{openai2025gpt5.6luna,
	title        = {OpenAI Models Documentation: GPT-5.6 Luna Specifications},
	author       = {{OpenAI}},
	year         = {2026},
	url          = {https://developers.openai.com/api/docs/models/gpt-5.6-luna},
	note         = {Accessed: 2026-09-21},
	organization = {OpenAI}
}

@misc{nguyen2024directevaluationchainofthoughtmultihop,
	title={Direct Evaluation of Chain-of-Thought in Multi-hop Reasoning with Knowledge Graphs}, 
	author={Minh-Vuong Nguyen and Linhao Luo and Fatemeh Shiri and Dinh Phung and Yuan-Fang Li and Thuy-Trang Vu and Gholamreza Haffari},
	year={2024},
	eprint={2402.11199},
	archivePrefix={arXiv},
	primaryClass={cs.CL},
	url={https://arxiv.org/abs/2402.11199}, 
}

@misc{rajpurkar2016squad100000questionsmachine,
	title={SQuAD: 100,000+ Questions for Machine Comprehension of Text}, 
	author={Pranav Rajpurkar and Jian Zhang and Konstantin Lopyrev and Percy Liang},
	year={2016},
	eprint={1606.05250},
	archivePrefix={arXiv},
	primaryClass={cs.CL},
	url={https://arxiv.org/abs/1606.05250}, 
}

@misc{brehme2026evaluatingmultihopreasoningrag,
	title={Evaluating Multi-Hop Reasoning in RAG Systems: A Comparison of LLM-Based Retriever Evaluation Strategies}, 
	author={Lorenz Brehme and Thomas Ströhle and Ruth Breu},
	year={2026},
	eprint={2604.18234},
	archivePrefix={arXiv},
	primaryClass={cs.IR},
	url={https://arxiv.org/abs/2604.18234}, 
}

@misc{singh2024evaluationdatacontaminationllms,
	title={Evaluation data contamination in LLMs: how do we measure it and (when) does it matter?}, 
	author={Aaditya K. Singh and Muhammed Yusuf Kocyigit and Andrew Poulton and David Esiobu and Maria Lomeli and Gergely Szilvasy and Dieuwke Hupkes},
	year={2024},
	eprint={2411.03923},
	archivePrefix={arXiv},
	primaryClass={cs.CL},
	url={https://arxiv.org/abs/2411.03923}, 
}

@misc{katsis2025mtragmultiturnconversationalbenchmark,
	title={MTRAG: A Multi-Turn Conversational Benchmark for Evaluating Retrieval-Augmented Generation Systems}, 
	author={Yannis Katsis and Sara Rosenthal and Kshitij Fadnis and Chulaka Gunasekara and Young-Suk Lee and Lucian Popa and Vraj Shah and Huaiyu Zhu and Danish Contractor and Marina Danilevsky},
	year={2025},
	eprint={2501.03468},
	archivePrefix={arXiv},
	primaryClass={cs.CL},
	url={https://arxiv.org/abs/2501.03468}, 
}

@misc{liu2026semanticcachinglowcostllm,
	title={Semantic Caching for Low-Cost LLM Serving: From Offline Learning to Online Adaptation}, 
	author={Xutong Liu and Baran Atalar and Xiangxiang Dai and Jinhang Zuo and Siwei Wang and John C. S. Lui and Wei Chen and Carlee Joe-Wong},
	year={2026},
	eprint={2508.07675},
	archivePrefix={arXiv},
	primaryClass={cs.LG},
	url={https://arxiv.org/abs/2508.07675}, 
}

@misc{gao_retrieval-augmented_2024,
	title = {Retrieval-Augmented Generation for Large Language Models: A Survey},
	url = {http://arxiv.org/abs/2312.10997},
	doi = {10.48550/arXiv.2312.10997},
	shorttitle = {Retrieval-Augmented Generation for Large Language Models},
	number = {{arXiv}:2312.10997},
	publisher = {{arXiv}},
	author = {Gao, Yunfan and Xiong, Yun and Gao, Xinyu and Jia, Kangxiang and Pan, Jinliu and Bi, Yuxi and Dai, Yi and Sun, Jiawei and Wang, Meng and Wang, Haofen},
	urldate = {2025-01-14},
	date = {2024-03-27},
    year ={2024},
	eprinttype = {arxiv},
}

@misc{park_dochop-qa_2025,
	title = {{DocHop}-{QA}: {Towards} {Multi}-{Hop} {Reasoning} over {Multimodal} {Document} {Collections}},
	shorttitle = {{DocHop}-{QA}},
	url = {http://arxiv.org/abs/2508.15851},
	doi = {10.48550/arXiv.2508.15851},
	urldate = {2026-01-12},
	publisher = {arXiv},
	author = {Park, Jiwon and Pyeon, Seohyun and Kim, Jinwoo and Cabal, Rina Carines and Ding, Yihao and Han, Soyeon Caren},
	month = aug,
	year = {2025},
	note = {arXiv:2508.15851 [cs]},
}

@misc{rosenthal2024clapnq,
      title={CLAPNQ: Cohesive Long-form Answers from Passages in Natural Questions for RAG systems}, 
      author={Sara Rosenthal and Avirup Sil and Radu Florian and Salim Roukos},
      year={2024},
      eprint={2404.02103},
      archivePrefix={arXiv},
      primaryClass={cs.CL}
}

@misc{noauthor_embeddings_nodate,
    author  = {{Google}},
  title   = {Embeddings},
  year    = {2026},
  url     = {https://ai.google.dev/gemini-api/docs/embeddings},
  urldate = {2026-08-20},
    note         = {Accessed: 2026-04-22},
	organization = {Google LLC}
}

@misc{noauthor_baaibge-small-en-v15_2026,
    title={C-Pack: Packaged Resources To Advance General Chinese Embedding}, 
      author={Shitao Xiao and Zheng Liu and Peitian Zhang and Niklas Muennighoff},
      year={2023},
      eprint={2309.07597},
      archivePrefix={arXiv},
      primaryClass={cs.CL}
}

@misc{noauthor_sentence-transformersall-minilm-l6-v2_2024,
	  title = "Making Monolingual Sentence Embeddings Multilingual using Knowledge Distillation",
    author = "Reimers, Nils and Gurevych, Iryna",
    booktitle = "Proceedings of the 2020 Conference on Empirical Methods in Natural Language Processing",
    month = "11",
    year = "2020",
    publisher = "Association for Computational Linguistics",
    url = "https://arxiv.org/abs/2004.09813",
}

@misc{noauthor_modellkarte_nodate,
 author  = {{Google DeepMind}},
  title   = {Gemma 4 Model Card},
  year    = {2026},
  url     = {https://ai.google.dev/gemma/docs/core/model_card_4},
  urldate = {2026-05-07}
}

@manual{noauthor_langchain_nodate,
  author    = {{LangChain}},
  title     = {{LangChain}},
  url       = {https://www.langchain.com/},
  urldate   = {2025-09-17},
    year = {2025},
}

@misc{lee2024multidocumentgroundedmultiturnsynthetic,
      title={Multi-Document Grounded Multi-Turn Synthetic Dialog Generation}, 
      author={Young-Suk Lee and Chulaka Gunasekara and Danish Contractor and Ramón Fernandez Astudillo and Radu Florian},
      year={2024},
      eprint={2409.11500},
      archivePrefix={arXiv},
      primaryClass={cs.CL},
      url={https://arxiv.org/abs/2409.11500}, 
}

@misc{k5nextK5NextClevere,
	author = {{Kufgem GmbH}},
	title = {k5\textbar {N}ext - {D}as clevere {G}emeindemanagement --- k5next.at},
	howpublished = {\url{https://www.k5next.at/}},
	year = {2026},
	note = {[Accessed 20-08-2026]},
}

@inproceedings{kukreja_performance_2024,
	title = {Performance Evaluation of Vector Embeddings with Retrieval-Augmented Generation},
	url = {https://ieeexplore.ieee.org/document/10603291},
	doi = {10.1109/ICCCS61882.2024.10603291},
	eventtitle = {2024 9th International Conference on Computer and Communication Systems ({ICCCS})},
	pages = {333--340},
	booktitle = {2024 9th International Conference on Computer and Communication Systems ({ICCCS})},
	author = {Kukreja, Sanjay and Kumar, Tarun and Bharate, Vishal and Purohit, Amit and Dasgupta, Abhijit and Guha, Debashis},
	urldate = {2024-10-25},
	date = {2024-04},
    year = {2024},
}

@article{lyu_crud-rag_2025,
author = {Lyu, Yuanjie and Li, Zhiyu and Niu, Simin and Xiong, Feiyu and Tang, Bo and Wang, Wenjin and Wu, Hao and Liu, Huanyong and Xu, Tong and Chen, Enhong},
title = {CRUD-RAG: A Comprehensive Chinese Benchmark for Retrieval-Augmented Generation of Large Language Models},
year = {2025},
issue_date = {March 2025},
publisher = {Association for Computing Machinery},
address = {New York, NY, USA},
volume = {43},
number = {2},
issn = {1046-8188},
url = {https://doi.org/10.1145/3701228},
doi = {10.1145/3701228},
journal = {ACM Trans. Inf. Syst.},
month = jan,
articleno = {41},
numpages = {32}
}

@inproceedings{schimanski-etal-2024-climretrieve,
    title = "{C}lim{R}etrieve: A Benchmarking Dataset for Information Retrieval from Corporate Climate Disclosures",
    author = "Schimanski, Tobias  and
      Ni, Jingwei  and
      Mart{\'i}n, Roberto Spacey  and
      Ranger, Nicola  and
      Leippold, Markus",
    editor = "Al-Onaizan, Yaser  and
      Bansal, Mohit  and
      Chen, Yun-Nung",
    booktitle = "Proceedings of the 2024 Conference on Empirical Methods in Natural Language Processing",
    month = nov,
    year = "2024",
    address = "Miami, Florida, USA",
    publisher = "Association for Computational Linguistics",
    url = "https://aclanthology.org/2024.emnlp-main.969/",
    doi = "10.18653/v1/2024.emnlp-main.969",
    pages = "17509--17524"
}

@misc{brehme_automating_2026,
	title = {Automating {Multi}-{Hop} {RAG} {Evaluation} via {TRIAD}: {From} {Context} {Extraction} to {Validated} {Dataset} {Generation}},
	shorttitle = {Automating {Multi}-{Hop} {RAG} {Evaluation} via {TRIAD}},
	url = {http://arxiv.org/abs/2608.21558},
	doi = {10.48550/arXiv.2608.21558},
	urldate = {2026-11-16},
	publisher = {arXiv},
	author = {Brehme, Lorenz and Jatowt, Adam},
	month = aug,
	year = {2026},
	note = {arXiv:2608.21558 [cs.CL]},
}

\section*{\uppercase{Appendix}}
\subsection*{Results}
\begin{table*}[ht]
\centering
\caption{Underlying Data for Figure \ref{fig:six}: Comparison of RAG modifications with the baseline (complete history, context list size = 10; cf. Table \ref{tab:consistency}, 5 turns).}
\label{tab:rag_modifcation}
\setlength{\tabcolsep}{4pt}
\renewcommand{\arraystretch}{0.9}
\small
\resizebox{\textwidth}{!}{%
\begin{tabular}{llcccccc}
\toprule
Modification
& Approach
& Context Retention
& Forgetfulness
& Correctness
& Faithfulness
& Context Precision
& Context Recall \\
\midrule

\multicolumn{8}{c}{\textbf{Consistency Baseline}} \\
\midrule
\multirow{2}{*}{MH-RAG-DIVE (All)}
& Min
& 0.7900 & 0.2100 & 0.6000 & 0.6996 & 0.5597 & 0.6390 \\
& Max
& 0.9300 & 0.4100 & 0.6560 & 0.7304 & 0.6115 & 0.6875 \\
\midrule
\multirow{2}{*}{MH-RAG-DIVE (Roots)}
& Min
& -- & -- & 0.6500 & 0.6647 & 0.4785 & 0.6883 \\
& Max
& -- & -- & 0.7000 & 0.7308 & 0.5184 & 0.7350 \\
\midrule

\multicolumn{8}{c}{\textbf{Baseline Evaluation of RAG-DIVE and ClapNQ}} \\
\midrule
RAG-DIVE
&
& 0.8800 & 0.4100 & 0.6520 & 0.7765 & 0.6285 & 0.6320 \\
ClapNQ
&
& -- & -- & 0.8567 & 0.8706 & 0.8989 & 0.8464 \\
\midrule

\multicolumn{8}{c}{\textbf{Embedding Model Modification}} \\
\midrule

\multirow{4}{*}{BAAI/bge-small-en-v1.5}
& MH-RAG-DIVE (All)
& \textbf{0.7300*} & \textbf{0.4900*} & \textbf{0.4960*}
& \textbf{0.5822*} & \textbf{0.4302*} & \textbf{0.5018*} \\
& MH-RAG-Dive (Roots) & -- & --  & \textbf{0.4800*} & \textbf{0.5247*} & \textbf{0.4052*} & \textbf{0.5950*} \\
& RAG-DIVE
& 0.8600 & 0.4800 & 0.4760 & 0.6460 & 0.4866 & 0.5178 \\
& ClapNQ
& -- & -- & 0.7733 & 0.9135 & 0.7854 & 0.7319 \\
\midrule

\multirow{4}{*}{sentence-transformers/all-MiniLM-L6-v2}
& MH-RAG-DIVE (All)
& \textbf{0.7600*} & 0.4100 & \textbf{0.4320*}
& \textbf{0.5186*} & \textbf{0.4233*} & \textbf{0.4770*} \\
& MH-RAG-Dive (Roots) & -- & -- & \textbf{0.2700*} & \textbf{0.3463*} & \textbf{0.3686*} & \textbf{0.5017*} \\
& RAG-DIVE
& 0.8200 & 0.5100 & 0.5520 & 0.6909 & 0.4894 & 0.5244 \\
& ClapNQ
& -- & -- & 0.7300 & 0.9050 & 0.7568 & 0.6760 \\
\midrule

\multicolumn{8}{c}{\textbf{Number of Contexts Modification}} \\
\midrule

\multirow{4}{*}{1 Context}
& MH-RAG-DIVE (All)
& \textbf{0.6800*} & \textbf{0.4200*} & \textbf{0.3080*}
& \textbf{0.3731*} & \textbf{0.5380*} & \textbf{0.3873*} \\
& MH-RAG-Dive (Roots) & -- & -- & \textbf{0.0500*} & \textbf{0.0958*} & \textbf{0.4200*} & \textbf{0.2650*} \\
& RAG-DIVE
& 0.8182 & 0.5354 & 0.4525 & 0.6434 & 0.6222 & 0.4116 \\
& ClapNQ
& -- & -- & 0.7200 & 0.9142 & 0.8867 & 0.4083 \\
\midrule

\multirow{4}{*}{8 Contexts}
& MH-RAG-DIVE (All)
& 0.8700 & 0.2300 & \textbf{0.6840*}
& \textbf{0.7915*} & 0.5717 & \textbf{0.7523*} \\
& MH-RAG-Dive (Roots) & -- & -- & \textbf{0.7800*} & \textbf{0.7999*} & \textbf{0.5303*} & \textbf{0.8067*} \\
& RAG-DIVE
& 0.8500 & 0.2800 & 0.6980 & 0.7990 & 0.6083 & 0.7142 \\
& ClapNQ
& -- & -- & 0.8633 & 0.9350 & 0.8512 & 0.8807 \\
\midrule

\multicolumn{8}{c}{\textbf{Agentic Modification}} \\
\midrule

\multirow{4}{*}{Agentic}
& MH-RAG-DIVE (All)
& \textbf{0.9700*} & \textbf{0.0500*} & \textbf{0.7960*}
& \textbf{0.8627*} & 0.5904 & \textbf{0.7307*} \\
& MH-RAG-Dive (Roots) & -- & -- & \textbf{0.8700*} & \textbf{0.8956*} & \textbf{0.4633*} & \textbf{0.8217*}\\
& RAG-DIVE
& 1.0000 & 0.0300 & 0.8000 & 0.8788 & 0.6563 & 0.6970 \\
& ClapNQ
& -- & -- & 0.8963 & 0.9325 & 0.8762 & 0.8217 \\
\bottomrule
\end{tabular}%
}

\vspace{2pt}
\begin{minipage}{\textwidth}
\footnotesize
\textit{*} Values outside the min--max range of the consistency baseline.
\end{minipage}
\end{table*}

\label{human_validation}
\begin{table}[ht]
    \centering
    \caption{Human Validation (sample size: 100 turns)}
    \begin{tabular}{p{0.2cm}p{0.6cm}p{0.8cm}p{0.8cm}p{0.8cm}p{0.6cm}p{0.8cm}}
    \toprule
        & correct & answer-able & plaus-ible & multi-hop & di-verse & co-herent  \\
         \midrule
        T1 & 93\% & 94\% & 91\% &  95\% &95\% & 90\% \\
        T2 & 93\% & 93\% & 89\% &  95\% &90\% & 90\% \\
        \bottomrule
    \end{tabular}
    \label{tab:data_quality}
\end{table}
\begin{table}[ht]
\centering
\caption{Evaluation Consistency: Overall vs. Multi-Hop Roots (n=10)}
\label{tab:consistency_full}
\begin{tabular}{lcccc}
\toprule
Metric & Mean & Std & Min & Max \\
\midrule
\multicolumn{5}{c}{\textbf{All Turns (Complete Dialogues)}} \\
\midrule
Context Retention  & 0.87    & 0.04    & 0.79    & 0.93    \\
Forgetfulness      & 0.30    & 0.05    & 0.21    & 0.41    \\
Correctness        & 0.63 & 0.02 & 0.60 & 0.66 \\
Faithfulness       & 0.71 & 0.01 & 0.7 & 0.73 \\
Context Precision  & 0.59 & 0.02 & 0.56 & 0.61 \\
Context Recall     & 0.67 & 0.01 & 0.64 & 0.69 \\
\midrule
\multicolumn{5}{c}{\textbf{Multi-Hop Roots Only (Turn 0)}} \\
\midrule
Correctness        & 0.68 & 0.02 & 0.65 & 0.70 \\
Faithfulness       & 0.69 & 0.02 & 0.66 & 0.73 \\
Context Precision  & 0.50 & 0.01 & 0.48 & 0.52 \\
Context Recall     & 0.71 & 0.02 & 0.69 & 0.74 \\
\bottomrule
\end{tabular}
\end{table}

\end{document}